\documentclass[10pt,twocolumn,letterpaper]{article}

\usepackage{times}
\usepackage{epsfig}
\usepackage{graphicx}
\usepackage{amsmath}
\usepackage{amssymb}
\usepackage{enumerate}
\usepackage{booktabs}
\usepackage{multirow}
\usepackage{xcolor}
\usepackage{color, colortbl}
\definecolor{Gray}{gray}{0.9}
\usepackage{adjustbox}
\usepackage{caption}
\usepackage{float}
\usepackage{subfigure}
\usepackage[bottom=1in,top=1in,left=0.8in,right=0.8in]{geometry}
\usepackage{abstract}

\usepackage[pagebackref=true,breaklinks=true,colorlinks,bookmarks=false]{hyperref}

\newcommand{\vx}{{\bf x}}

\newcommand{\vz}{{\bf z}}

\begin{document}

\title{\bf Few-Shot Image Classification Along Sparse Graphs} 

\author{Joseph F Comer\and Philip L Jacobson\and Heiko Hoffmann \\
HRL Laboratories, LLC\\
3011 Malibu Canyon Rd, Malibu, CA 90265\\
{\tt\small jfcomer@hrl.com, pljacobson@hrl.com, hhoffmann@hrl.com}
}
\date{}

\maketitle
\thispagestyle{empty}

\begin{abstract}
    Few-shot learning remains a challenging problem, with unsatisfactory 1-shot accuracies for most real-world data. Here, we present a different perspective for data distributions in the feature space of a deep network and show how to exploit it for few-shot learning. First, we observe that nearest neighbors in the feature space are with high probability members of the same class while generally two random points from one class are not much closer to each other than points from different classes. This observation suggests that classes in feature space form sparse, loosely connected graphs instead of dense clusters. To exploit this property, we propose using a small amount of label propagation into the unlabeled space and then using a kernel PCA reconstruction error as decision boundary for the feature-space data distribution of each class. Using this method, which we call ``K-Prop,'' we demonstrate largely improved few-shot learning performances (e.g., 83\% accuracy for 1-shot 5-way classification on the RESISC45 satellite-images dataset) for datasets for which a backbone network can be trained with high within-class nearest-neighbor probabilities. We demonstrate this relationship using six different datasets.
\end{abstract}

\section{Introduction}

Learning from few labeled examples, or ``few-shot" learning, is needed for applications where labels are expensive or hard to obtain or where adapting to new data has to be fast. But few-shot learning remains a challenging problem. Particularly, with only 1 to 5 labels per class, classification accuracies are typically low on real-world data~\cite{ibm}. 

The simplest approach to few-shot learning is to adapt (fine-tune) a pre-trained network to a new target dataset based on the small set of available labeled data~\cite{baseline++}. Either the weights of the entire network are adapted or only the final classification layer. In the latter case, the network is typically split into two parts: a backbone network, consisting, e.g., of multiple convolutional and pooling layers and a head consisting of a multi-layer perception or a linear mapping. Here, the backbone maps the input data into a so-called feature space. For few-shot learning the weights of the backbone are typically frozen while the weights of the head are adapted. With comprehensive pre-training of the backbone (on a meta training set),  training a linear classifier in the feature space can provide decent results~\cite{goodembedding}, though the 1-shot results are typically in the best case in the 60-70\% accuracy regime for 5-way classification.

\begin{figure}
\begin{center}
\includegraphics[width=3.3in]{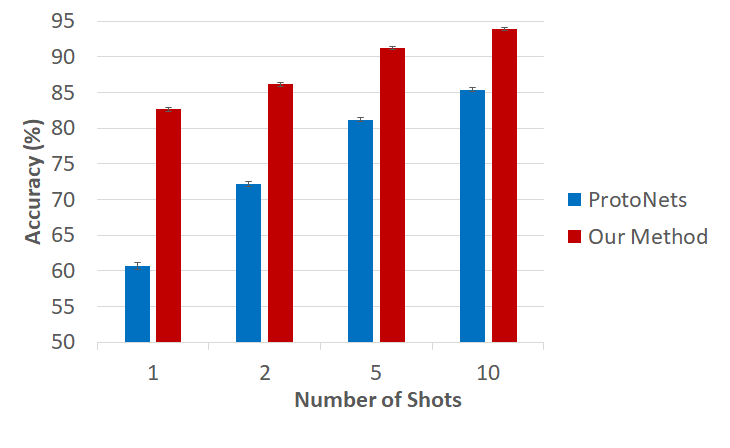}
\end{center}
\vspace{-0.2in}
   \caption{Comparing the accuracy of our method, K-Prop, with ProtoNets on the RESISC45 dataset for 5-way classification (mean $\pm$ SE).}
\label{fig:protonetcomp}
\end{figure}

\begin{figure*}[ht]
\begin{center}
\includegraphics[width=6in]{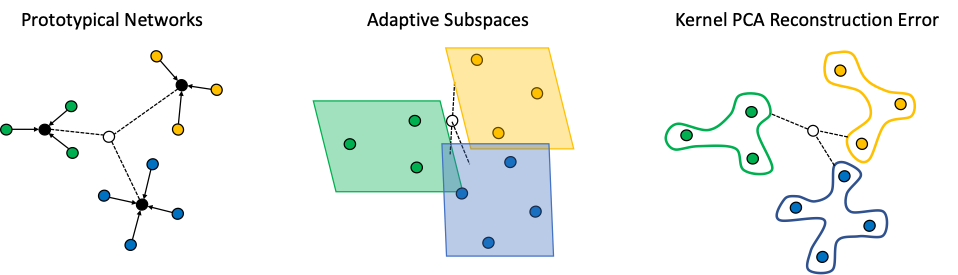}
\end{center}
   \caption{Methods for classification in feature space: ProtoNets~\cite{protonets}, Subspaces~\cite{subspaces}, and kernel PCA.}
\label{fig:featureSpace}
\end{figure*}

As alternatives to a linear classifier baseline, various methods have been suggested, broadly falling into three categories: initialization-based, metric-based, and generative methods. Initialization-based methods, such as MAML \cite{MAML,iMAML}, search for a good initialization of the model weights for which good performance on downstream tasks can be achieved with few labeled examples and in few training epochs. Metric-based methods work by replacing the linear classifier with a more sophisticated way of comparing distances between unseen data and the few labeled points. Some examples include ProtoNets~\cite{protonets}, which compute distances to prototypical references points for each class, and Adaptive Subspaces~\cite{subspaces}, which compute distances to subspaces fitted to the data distribution of a class in feature space (Fig. \ref{fig:featureSpace}). These methods do not necessarily provide an improvement over a linear classifier, which can, e.g., beat ProtoNets if having a well trained backbone~\cite{goodembedding}. Lastly, generative methods work by using a generator to augment the original data with novel data for training \cite{MetaGAN}. MetaGAN uses an existing few-shot learning method and boosts its performance by generating additional data, though the reported improvements have been only by a few percent~\cite{MetaGAN}.

Here, we provide an alternative to these approaches for classifying data in the feature space. As a backbone network, we consider either a pretrained network or a self-supervised trained network, which is trained on the unlabeled target data. Here, we assume that we have access to a large amount of unlabeled data in the target domain, while the labeled data are scant.

As a background, we observed that the pairwise distances between images in feature space are fairly similar to each other (most are within $\pm 50$\% of the mean distance). That is, we cannot expect data points of one class be clustered together in Euclidean space. In addition, we observed that for some datasets, the nearest neighbors in feature space belong with high probability ($>90$\%) to the same class. These two properties suggest a data distribution of a class that resembles a loosely connected sparse graph, and the graphs of different classes are tightly intermingled.

Based on these insights, we construct a few-shot learning method as follows: given a few labeled points in feature space, we first propagate the labels to nearby unlabeled points over nearest-neighbor links. Second, given the resulting data-point distribution, we compute the kernel principal component analysis (kernel PCA) reconstruction error~\cite{KPCArecerr} as a distance measure of a test point to each class. The hypersurfaces of equal kernel PCA reconstruction error have been shown to follow the shape of any data-point distribution~\cite{KPCArecerr}. 

Our experiments show that this new method outperforms state-of-the-art few-shot learning methods, like ProtoNets and Adaptive Subspaces, particularly, for 1-shot learning on certain data sets. We found that these datasets in combination with a backbone network share a common property: in feature space, the nearest neighbors are with high probability part of the same class.   

\vspace{5mm}
\noindent {\bf Contributions.} In summary, we make the following three main contributions:

\begin{enumerate}[i]
  \item Provide a new perspective on how to interpret data in the feature space based on our observation of pairwise data-point distances.
  \item Introduce a new method, K-Prop, combining self-supervised learning, label propagation, and kernel PCA for few-shot learning.
  \item Show a relationship between high few-shot learning performance with our method and data-point distances in feature space.
\end{enumerate}

The remainder of this article is organized as follows: Section 2 describes related work; Section 3 provides the background for our new method including our observation of pairwise data-point distances and nearest-neighbor relationships; Section 4 describes our proposed method; Section 5 our experiments with the corresponding results in Section 6. Finally, Section 7 concludes with a summary, discussion of limits and potential risks to society, and outlook for future work.

\section{Related Work}

For few-shot learning, a subset of methods deals with different means for mapping from the feature space onto the image classes. This mapping is generally much less complex than the mapping from the image space onto the feature space, for which usually deep multi-layer convolutional neural networks are used. This reduced complexity allows for adaptation to a target dataset with only a few labels. Examples of this approach are  ProtoNets~\cite{protonets} and Adaptive Subspaces~\cite{subspaces}. 

ProtoNets were introduced as an improvement over MatchingNets~\cite{matchingnets}, which is a popular few-shot learning technique, and Adaptive Subspaces were introduced as an improvement over ProtoNets~\cite{subspaces}. From a certain angle, our approach could be viewed as an extension of Adaptive Subspaces to non-linear subspaces, which we describe with kernel PCA. So, there is a natural progression of methods from the reference points in ProtoNets over subspaces to kernel PCA (Fig. \ref{fig:featureSpace}).

Label propagation has been suggested before. For example, Zhou et al.~\cite{labelprop_diffusion} introduced label propagation through diffusion in a semi-supervised learning setting if the data manifold is sufficiently smooth. More recently, authors studied label propagation in feature space: Iscen et al.~\cite{labelprop_CVPR19} construct a k-nearest-neighborhood graph and propagate labels with a diffusion process. Liu et al.~\cite{labelprop-learn} construct a neighborhood graph with a Gaussian similarity matrix and learn the parameters for label propagation in a meta-learning setting. Benato et al.~\cite{labelprop-tSNE} map from the feature space onto a t-SNE-generated 2-dimensional plane before propagating labels.

None of these works mentioned the property that we observed: that nearest neighbors can be with high probability within the same class for certain datasets and backbones.

\section{Background}

In this section, we provide the background for our new method: 1) self-supervised learning of a backbone network, 2) our observation of pairwise data-point distances, and 3) the kernel PCA reconstruction error.

\subsection{Self-supervised learning}

As unlabeled data are often less expensive to obtain, one common approach to few-shot learning is to use so-called self-supervision, wherein a proxy task is employed to pretrain a backbone network to produce features which can be leveraged for the downstream few-shot classification task. 
Using self-supervised learning to train a backbone network has been shown to rival {\em supervised} training based on linear-classifier accuracy on the trained features for certain datasets~\cite{simCLR,EsViT}.

Li et al.~\cite{EsViT} train various architectures in the multistage vision transformers (ViT) family using a self-supervision scheme which builds off of DINO \cite{DINO}. ViT architectures work by first splitting an image into a regular grid of non-overlapping patches, flattening and (optionally) projecting the patches, and then performing sparse multi-head attention on the collection of patches. 

In DINO, an exponential moving average ViT teacher network and a student network of the same architecture are fed different augmentations (views) of the full image, and the view-level features produced by each network are then fed to a shared prediction head which maximizes agreement between the feature representations of the two views. Li et al.~\cite{EsViT} use an additional region-level task which similarly enforces similarity of the top-level feature representations of the various patches by first matching each student feature to the most similar output feature of the same layer of the teacher network, and then using the mean similarity of the resulting collection of feature pairs as the loss (see \cite{EsViT} for full details).

\subsection{Pairwise data-point distances}

\begin{figure}
\begin{center}
\includegraphics[width=3.2in]{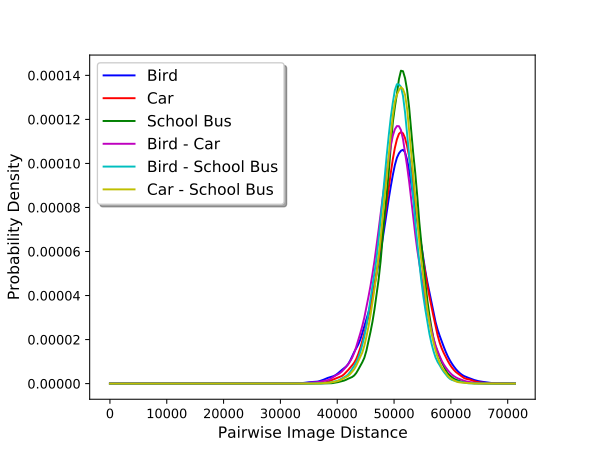}
\end{center}
\vspace{-0.1in}
   \caption{Pairwise distances (within and between classes) in the original image space for selected Imagenet classes.}
\label{fig:imageDist}
\end{figure}

We investigated the geometric structure of images, as embedded in the high-dimensional space where each pixel describes one dimension. For a selected subset of Imagenet~\cite{imagenet} classes, we computed pairwise distances between images within each class and between classes. As a result, in the original image space, the data points have almost the same distance to each other (Fig. \ref{fig:imageDist}).

\begin{figure}
\begin{center}
\includegraphics[width=3.2in]{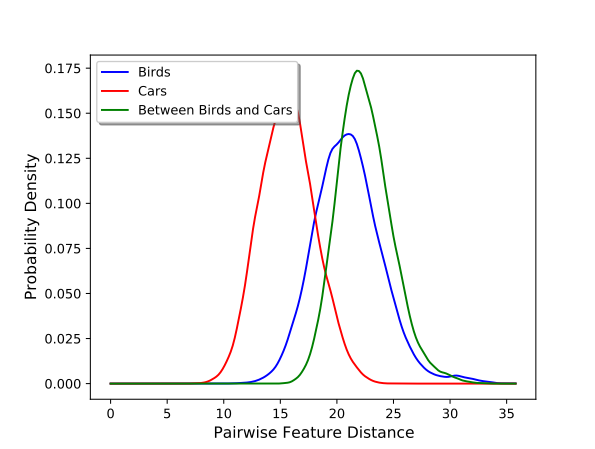}
\end{center}
\vspace{-0.1in}
   \caption{Pairwise distances in the 512-dimensional feature space of an Imagenet-pretrained Resnet18.}
\label{fig:resnet18Dist}
\end{figure}

Next, we mapped the images into the 512-dimensional feature space of a ResNet18 network, which was pre-trained on Imagenet. Here, 
the pairwise distances show more structure (Fig. \ref{fig:resnet18Dist}), but a large part of the data still have similar distances to each other. In addition, for the Imagenette dataset, we computed the pairwise distances in the 512-dimensional feature space of an Imagenet-pretrained ResNet18 network. For the intra-class pairwise distances, we observed a mean of $23.25 \pm 3.48$ SD ($n = 4.5*10^6$). For the inter-class distances, we observed $29.06 \pm 3.15$ (mean $\pm$ SD, $n = 4*10^7$). That is, the difference between inter and intra distances is relatively small, which matches the qualitative observation in Fig. \ref{fig:resnet18Dist}.

In addition, we evaluated the probability, $p_{NN}$, of a nearest neighbor in feature space being in the same class. We computed this probability for the Imagenet-pretrained ResNet18 features (Tab. \ref{tab:resnet18NNb}) as well as for the EsViT features (Tab. \ref{tab:esvitNNb}) for six different datasets (RESISC45, CUB, Imagenette, EuroSat, CropDisease, and Fungi - see also Experiments). For RESISC45, Imagenette, EuroSat, and CropDisease, we computed 100 trials, in each picking either 5 or 10 classes at random. In each trial, we computed the pairwise distances in feature space between all training data of the selected classes. For CUB and Fungi, we computed $1,000$ trials. 

As a result, in most cases, these probabilities, $p_{NN}$, were in the 90s\%. An outlier is CUB, for which $p_{NN}$ was low ($<60$\%) for the EsViT features.

\begin{table}[ht]
    \centering
    \begin{small}
    \begin{tabular}{c||c|c|c}
        \# of Classes & RESISC & CUB & Imagenette\\
        \hline\hline
        5 & $94.5 \pm 5.1$ & $90.0 \pm 11.2$ & $98.1 \pm 1.5$\\
        \hline
        10 & $89.5 \pm 8.2$ & $82.6 \pm 14.6$ & $96.7 \pm 2.5$\\
    \end{tabular}\\
    \begin{tabular}{c||c|c|c}
        \# of Classes & EuroSat & Crop & Fungi\\
        \hline\hline
        5 & $92.5 \pm 6.5$ & $97.7 \pm 4.6$ & $80.0 \pm 19.0$\\
        \hline
        10 & $86.4 \pm 9.8$ & $95.8 \pm 6.5$ & $70.9 \pm 20.7$\\
    \end{tabular}
    \end{small}
    \caption{Probability that nearest neighbor is within the same class for the ImageNet-pretrained {\bf Resnet18} features (mean $\pm$ SD).}
    \label{tab:resnet18NNb}
\end{table}

\begin{table}[ht]
    \centering
    \begin{small}
    \begin{tabular}{c||c|c|c}
        \# of Classes & RESISC & CUB & Imagenette\\
        \hline\hline
        5 & $98.7 \pm 2.1$ & $59.7 \pm 19.0$ & $92.0 \pm 4.0$ \\
        \hline
        10 & $97.1 \pm 3.3$ & $43.8 \pm 20.4$ & $87.8 \pm 6.8$ \\
    \end{tabular}\\
    \begin{tabular}{c||c|c|c}
        \# of Classes & EuroSat & Crop & Fungi\\
        \hline\hline
        5 & $98.0 \pm 1.5$ & $99.3 \pm 1.6$ & $88.7 \pm 13.3$ \\
        \hline
        10 & $96.3 \pm 2.2$ & $98.5 \pm 2.6$ & $83.2 \pm 15.5$ \\
    \end{tabular}
    \end{small}
    \caption{Probability that nearest neighbor is within the same class for the self-supervised trained {\bf EsViT} features (mean $\pm$ SD).}
    \label{tab:esvitNNb}
\end{table}

\begin{figure}
\begin{center}
\includegraphics[width=3.2in]{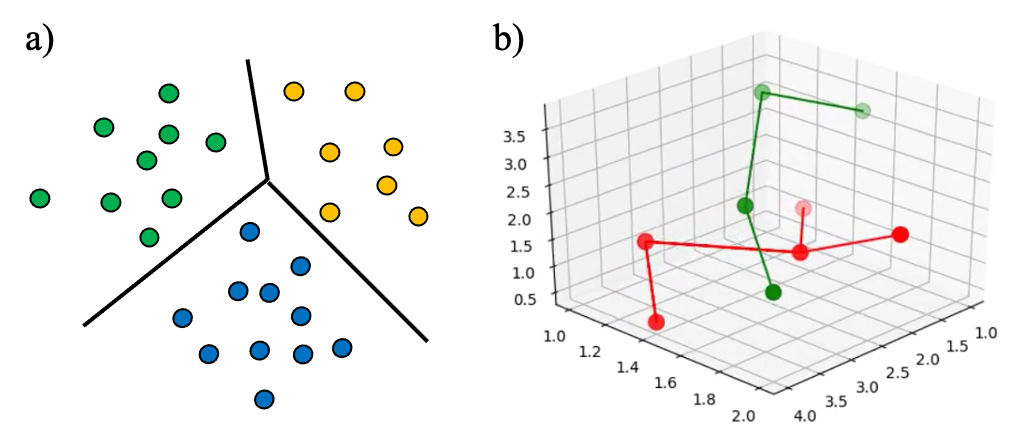}
\end{center}
   \caption{a) Common illustration of data point distributions in feature space, b) our interpretation: intermingled sparse graphs. Each color denotes a different class.}
\label{fig:illustration}
\end{figure}

Figure \ref{fig:illustration}a shows a common illustration of the data-point distribution in feature space. Based on our observations, we found this illustration to be misleading because the points of one class are not locally clustered together. This lack of clustering is evidenced by the small difference between intra-class and inter-class distances. So, instead, we propose that the data are distribution along sparse graphs, and the graphs between classes are intermingled (Fig. \ref{fig:illustration}b).

\subsection{Kernel Principle Component Analysis}

Kernel Principal Component Analysis is a kernelized version of the 
PCA algorithm, essentially, expanding the linear method to non-linear data distributions~\cite{KPCA}. Kernel PCA uses the so-called ``kernel trick,'' i.e., the PCA is computed in a high-dimensional (potentially, infinitely dimensional) space, into which all data points, $\{\bf{x}_i\}$, are mapped, without actually carrying out the mapping, ${\bf\Phi}(\bf{x}_i)$, into this space because the mapping appears only inside scalar products, and so, the scalar product can be replaced with a kernel function in the original space. A common kernel function is a Gaussian function,
\begin{equation}
k(\vx_i,\vx_j) = \exp(-||\vx_i-\vx_j||^2/(2\sigma^2))\,.   
\end{equation}
For this function, the corresponding mapping would actually be one into an infinite-dimensional space. Practically, however, the dimensionality is limited by the number of data points. 

Computing kernel PCA involves computing the kernel matrix $K_{ij} = k(\vx_i,\vx_j)$, transforming $K$ to account for the non-zero mean of $\{{\bf\Phi}(\bf{x}_i)\}$, and extracting a number, $q$, of eigenvectors corresponding to the $q$-largest eigenvalues \cite{KPCA,KPCArecerr}. Here, for few-shot learning, we are dealing with a low-dimensional kernel matrix, so the computational cost of the eigenvalue extraction is negligible. 

When computing kernel PCA for a non-linear data distribution, the corresponding reconstruction error (analogue to the reconstruction error in PCA) was introduced as a novelty-detection measure ~\cite{KPCArecerr}. 
For Gaussian kernel functions, it turned out that the equipotential curves/surfaces of the reconstruction error describe well the non-linear shape of a data-point distribution~\cite{KPCArecerr}. Thus, we use this reconstruction error to compare between different classes in features space. 

Let $\bf{z}$ be a vector in feature space and $\{\bf{x}_i\}$ be a distribution of vectors in feature space. Then, the reconstruction error of $\bf{z}$  is computed as follows~\cite{KPCArecerr},
\begin{eqnarray}\label{eq:recerr}
L_{RE}(\vz) &=& k(\vz,\vz) - \frac{2}{n}\sum_{i=1}^n k(\vz,\vx_i) \nonumber\\&+& \frac{1}{n^2} \sum_{i,j=1}^n k(\vx_i,\vx_j) - \sum_{l=1}^q f_l(\vz)^2\enspace,
\end{eqnarray}
where $f_l$ are the projections onto the principal components,
\begin{eqnarray}\label{eq:projection}
f_l(\vz) &=& \sum_{i=1}^n \alpha_i^l\, \Bigg[ k(\vz,\vx_i) - \frac{1}{n}\sum_{r=1}^n k(\vx_i,\vx_r)\nonumber\\
&-& \frac{1}{n}\sum_{r=1}^n k(\vz,\vx_r)+\frac{1}{n^2}\sum_{r,s=1}^n k(\vx_r,\vx_s)\Bigg]\,,
\end{eqnarray}
and $\alpha_i^l$ are the eigenvectors of $K$. When using a Gaussian kernel, we have two hyperparameters, the width, $\sigma$, and the number of principal components, $q$.

\section{Proposed Method}\label{sec:method}

\begin{figure}[h]
\begin{center}
\includegraphics[width=3.3in]{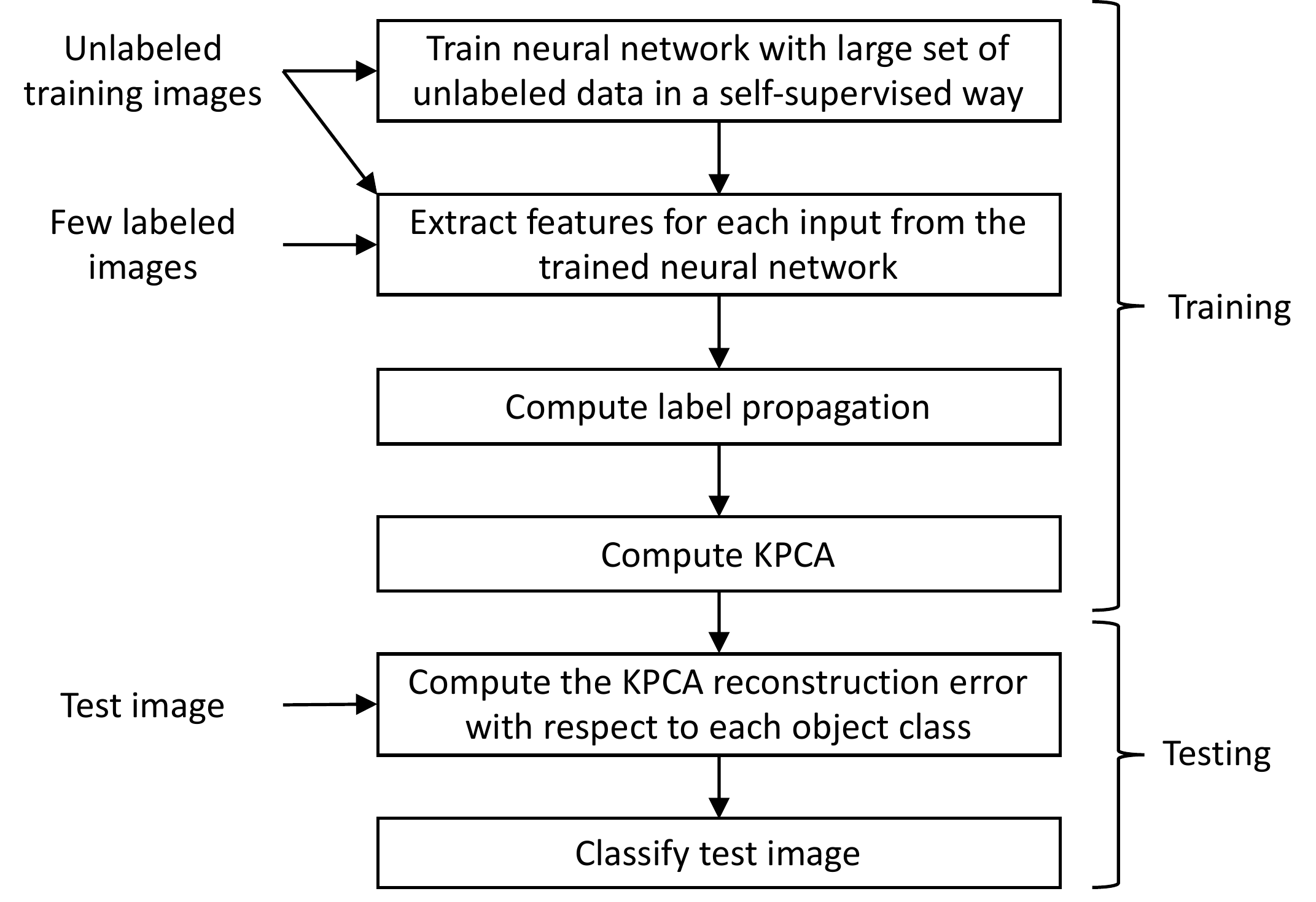}
\end{center}
\vspace{-0.2in}
   \caption{Process flow of our new method, {\em K-Prop}.}
\label{fig:method}
\end{figure}

Based on the above background, we propose a new few-shot learning method, {\em K-Prop}, using self-supervised learning, label propagation, and the kernel PCA reconstruction error (Fig. \ref{fig:method}). First, we exploit the fact that for many datasets, a good backbone, and thus a good mapping onto a feature space, can be obtained with self-supervised pre-training. Second, we exploit the nearest-neighbor within-class connections to artificially expand the number of labels with label propagation, and third, we exploit that the reconstruction error for kernel PCA can describe the sparse graph-like distribution of the few labeled points. In the following, we describe those three elements in more detail.

1) For self-supervised learning (SSL), we suggest the EsViT method \cite{EsViT}. We use SSL on the target training data (but not on the test data) to avoid cross-domain transfer loss. After training, we freeze the backbone parameters and weights.

2) For label propagation, we iteratively add a fixed number of extra labels. So, we propagate only into the neighborhood of the given labels, instead of diffusing into the entire unlabeled set. In each iteration step, we add only one unlabeled data point: in feature space, we find the point $\vx_j$ with the smallest Euclidean distance to any of the points $\vx_i$ in the set of labeled points, i.e.,

\begin{equation}
j = \mbox{argmin}_j \mbox{min}_i ||\vx_i - \vx_j||\,.
\end{equation}

\begin{figure}[h]
\begin{center}
\includegraphics[width=2.5in]{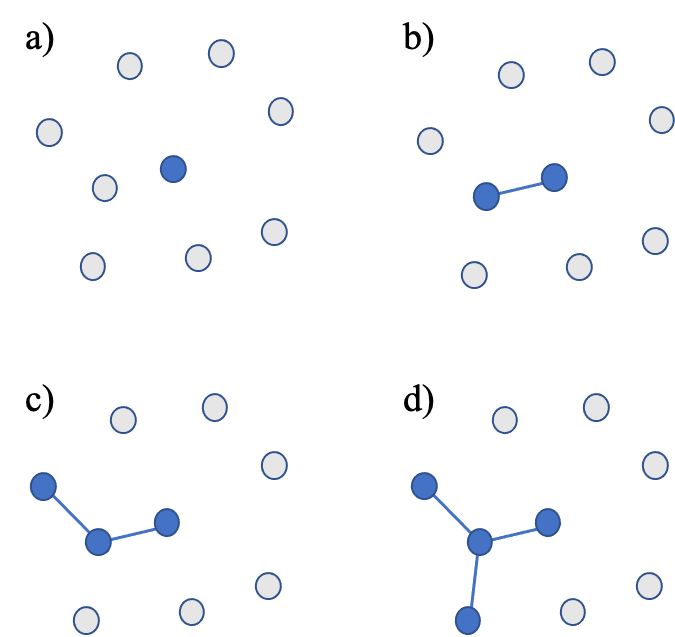}
\end{center}
\vspace{-0.1in}
   \caption{Label propagation: we iteratively add one unlabeled nearest neighbor at a time (a to d) building a graph of labeled points for each class.}
\label{fig:labelProp}
\end{figure}

The resulting $\vx_j$ is then added to the set of labeled points, and the iteration continues until a given number of points is added to the number of originally labeled points. Figure \ref{fig:labelProp} illustrates one example of this iterative process. Here, adding a nearest neighbor is shown as a link in a graph. For 1-shot learning, this process results in a sparse graph, and when starting with more labels, we generally end up with a multitude of graphs. 

3) We compute kernel PCA for each set of labeled feature points $\{\vx_i\}$ for each class separately. Here, the labeled data contain the extra labels from the label propagation. When presenting a new test image, we first compute its mapping into the feature space, $\vz$, and then compute the reconstruction error, $L_{RE}^c(\vz)$ for each class, $c$. We classify the test image based on the smallest reconstruction error, $\mbox{argmin}_c L_{RE}^c(\vz)$.


\begin{table*}[thb]
    \centering
    \begin{small}
    \textbf{RESISC}\\
    \begin{tabular}{c||c|c|c|c|c}
        \# of Shots & ProtoNets & MatchingNets & Subspaces & Ours (EsViT) & Ours (Resnet18) \\
        \hline\hline
        1 & $60.66 \pm 0.9$ & $60.21 \pm 0.9$ & $60.55 \pm 0.8$ & $\mathbf{83.18 \pm 0.6}$ & $71.15 \pm 0.7$\\
        \hline
        2 & $72.18 \pm 0.7$ & $69.59 \pm 0.7$ & $ 69.17 \pm 0.7$ & $\mathbf{87.06 \pm 0.5}$ & $76.52 \pm 0.6$\\
        \hline
        5 & $81.21 \pm 0.5$ & $75.75 \pm 0.5$ & $ 79.16 \pm 0.5$ & $\mathbf{92.08 \pm 0.4}$ & $84.84 \pm 0.5$\\
    \end{tabular}\\
    \textbf{Crop}\\
    \begin{tabular}{c||c|c|c|c|c}
        \# of Shots & ProtoNets & MatchingNets & Subspaces & Ours (EsViT) & Ours (Resnet18)\\
        \hline\hline
        1 & $77.40 \pm 0.8$ & $\mathbf{79.33 \pm 0.8}$ & $75.87 \pm 0.8$ & $78.11 \pm 0.8 $ & $78.53 \pm 0.7$\\
        \hline
        2 & $\mathbf{89.08 \pm 0.6}$ & $85.21 \pm 0.7$ & $ 81.41 \pm 0.6$ & $85.20 \pm 0.6 $ & $83.56 \pm 0.6$\\
        \hline
        5 & $91.38 \pm 0.4$ & $90.76 \pm 0.4$ & $ 91.95 \pm 0.5$ & $\mathbf{92.82 \pm 0.4}$ & $91.20 \pm 0.4$\\
    \end{tabular}\\
    \textbf{EuroSat}\\
    \begin{tabular}{c||c|c|c|c|c}
        \# of Shots & ProtoNets & MatchingNets & Subspaces & Ours (EsViT) & Ours (Resnet18)\\
        \hline\hline
        1 & $39.93 \pm 0.7$ & $37.40 \pm 0.7$ & $27.55 \pm 0.6$ & $\mathbf{77.20 \pm 0.5}$ & $67.90 \pm 0.7$\\
        \hline
        2 & $44.97 \pm 0.7$ & $39.57 \pm 0.5$ & $ 36.85 \pm 0.6$ & $\mathbf{83.40 \pm 0.4}$ & $74.54 \pm 0.6$\\
        \hline
        5 & $55.08 \pm 0.5$ & $44.86 \pm 0.6$ & $ 40.42 \pm 0.5$ & $\mathbf{90.44 \pm 0.3}$ & $82.86 \pm 0.4$\\
    \end{tabular}\\
    \textbf{Imagenette}\\
    \begin{tabular}{c||c|c|c|c|c}
        \# of Shots & ProtoNets & MatchingNets & Subspaces & Ours (EsViT) & Ours (Resnet18)\\
        \hline\hline
        1 & $31.57 \pm 0.6$ & $28.95 \pm 0.6$ & - & $\mathbf{74.73 \pm 0.6}$ & $91.47 \pm 0.4$ \\
        \hline
        2 & $30.99 \pm 0.5$ & $30.64 \pm 0.5$ & -& $\mathbf{81.12 \pm 0.4 }$ & $94.77 \pm 0.2$ \\
        \hline
        5 & $37.76 \pm 0.5$ & $36.98 \pm 0.5$ & - & $\mathbf{84.90 \pm 0.4}$ & $97.28 \pm 0.1$\\
    \end{tabular}\\
    \textbf{CUB}\\
    \begin{tabular}{c||c|c|c|c|c}
        \# of Shots & ProtoNets & MatchingNets & Subspaces & Ours (EsViT) & Ours (Resnet18)\\
        \hline\hline
        1 & $71.88 \pm 0.9$ & $72.36 \pm 0.9$ & $63.30 \pm 0.9$ & $40.79 \pm 0.7 $ & $\mathbf{81.08 \pm 0.8}$\\
        \hline
        2 & $81.44 \pm 0.6$ & $80.18 \pm 0.7$ & $68.38 \pm 0.8$ & $46.44 \pm 0.9 $ & $\mathbf{83.77 \pm 0.7}$ \\
        \hline
        5 & $87.42 \pm 0.5$ & $83.64 \pm 0.6$ & $78.25 \pm 0.6$ & $53.59 \pm 1.0$ & $\mathbf{89.39 \pm 0.6}$\\
    \end{tabular}\\
    \textbf{Fungi}\\
    \begin{tabular}{c||c|c|c|c|c}
        \# of Shots & ProtoNets & MatchingNets & Subspaces & Ours (EsViT) & Ours (Resnet18)\\
        \hline\hline
        1 & $58.87 \pm 0.9$ & $\mathbf{60.79 \pm 0.9}$ & $35.34 \pm 0.8$ & $47.26 \pm 1.1$ & $42.87 \pm 1.1$\\
        \hline
        2 & $\mathbf{74.65 \pm 0.8}$ & $74.21 \pm 0.7$ & $39.12 \pm 0.8$& $68.18 \pm 1.3 $ & $56.93 \pm 1.2$\\
        \hline
        5 & $\mathbf{82.13 \pm 0.6}$ & $80.64 \pm 0.8$ & $63.41 \pm 0.7$ & $74.29 \pm 1.2$ & $64.21 \pm 1.2$\\
    \end{tabular}\\
    \end{small}
   
    \caption{Comparing the performance of our method (with either EsViT or Resnet18 backbone) with ProtoNets, MatchingNets, and Adaptive Subspaces on the RESISC, CropDisease, Eurosat and Imagenette datasets (mean $\pm$ SE). We disregard the Imagenette results with Imagenet-pretrained Resnet18 when comparing with other methods because of unfair advantage. The Adpative Subspaces method failed to converge when training on Imagenette, and so the results are missing.}
    \label{tab:methodcomp}
\end{table*}

\section{Experiments}

We evaluate our method on six datasets and compare it against three other state-of-the-art methods and in ablation studies replacing key elements of K-Prop. Moreover, we consider features produced from backbones trained either by fully-supervised pretraining on a dissimilar source domain or by self-supervision with zero labels on the target domain. In the former case, we used a Resnet18 network with frozen weights, which has been pretrained on Imagenet-1k. In the latter case, we trained a multistage transformer architecture with the self-supervision scheme proposed for EsViT \cite{EsViT}.

For EsViT, we used the tiny sliding window architecture (Swin-T \cite{EsViT}). Images were divided into non-overlapping $16\times 16$ pixel patches and two additional (global) random crops of size $224\times 224$, all of which were subject to random transformations (augmentations) as described in \cite{SSViT}. We used a base learning rate of 0.0005 and cosine annealing, with a weight decay scaling linearly from 0.04 to 0.4 over 300 epochs. The network was trained for 300 epochs or until loss convergence.

For label propagation, we added 4 extra labels for 1-shot learning, 3 extra labels for 2-shot learning, and 2 extra labels for 5- and 10-shot learning. For 1-shot learning, the accuracies started to saturate by the 4th extra label, and too many extra labels will hurt because of pollution with wrong labels (see Appendix, Fig. \ref{fig:labels_added}).

For kernel PCA, we used a Gaussian kernel with width $\sigma = 16$ and $q = \lfloor 2k/3+1\rfloor$ principal components, where $k$ is the number of labels per class (before label propagation).

We compare our method against ProtoNets~\cite{protonets}, MatchingNets~\cite{matchingnets}, and Adaptive Subspaces~\cite{subspaces}. Each of the methods was implemented with a Resnet18 backbone, while otherwise following the training and augmentation routines in \cite{baseline++,subspaces}. Since these methods require meta learning to adapt parameters, we trained all three models using the labels for half of the classes in the dataset, while the other half was used to evaluate the few-shot learning performance. In contrast, using EsViT, our method did not use any labels (apart from few-shot) but used all classes of the unlabeled training set.

For our target datasets, we used Imagenette~\cite{imagenette}, RESISC45~\cite{resisc}, CropDisease~\cite{crop}, EuroSat~\cite{eurosat}, Fungi \cite{fungi}, and CUB \cite{CUB}. Imagenette is distributed under the Apache license, CropDisease under Creative Commons 1.0 Universal, Fungi under MIT, and CUB under Attribution 4.0 International. The licenses for RESISC45 and EuroSat are unknown. For all datasets, we used the default training/test data split.

For each dataset and for each pre-training scheme, we evaluate 5-way, k-shot performance using $k=1, 2, 5$. For each $k$, we randomly generate $1,000$ tasks by sampling, for each task, 5 uniformly random classes from the test set and, then average the classification accuracies over all tasks. The backbone weights were frozen after pretraining. For each task, we evaluate our method using either the EsViT or Resnet18 backbone, compare against other methods and do ablation studies with our method. For these ablation studies, we replaced kernel PCA with a linear classifier and tested the linear classifier either with or without label propagation. 

\section{Results}

On several datasets, our new method, K-Prop, outperformed state-of-the-art methods (Tab. \ref{tab:methodcomp}). For example, for 1-shot learning on RESISC45, K-Prop had a 83\% accuracy compared to 61\% for ProtoNets (Fig. \ref{fig:protonetcomp}). As another comparison, the best known 1-shot 5-way accuracy reported in the literature for RESISC45 with a Resnet18 backbone is $64.6$\%~\cite{ibm}. 

\begin{figure}[h]
\vspace{-0.1in}
\begin{center}
\includegraphics[width=3in]{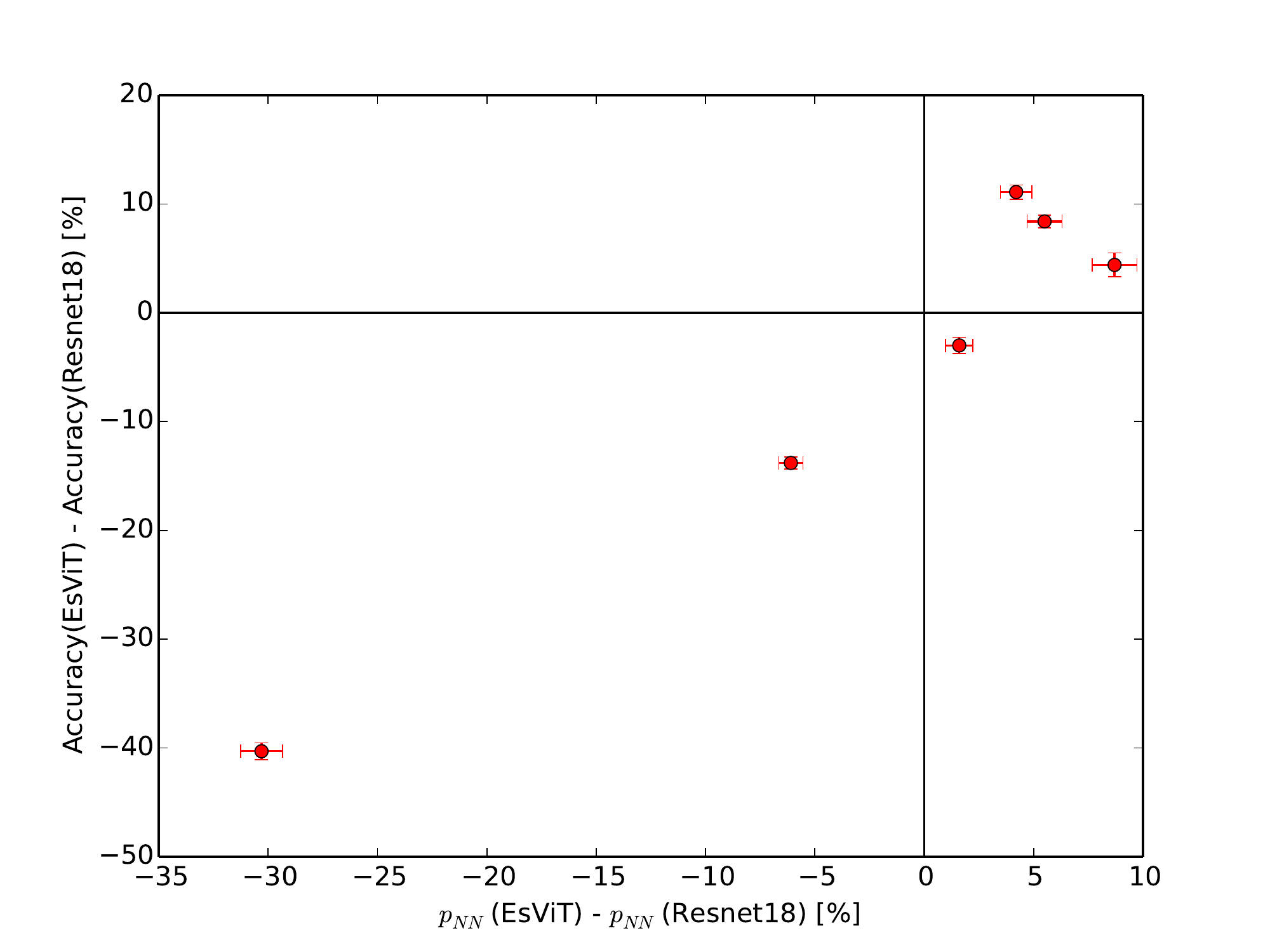}
\end{center}
\vspace{-0.2in}
   \caption{Comparison of 1-shot accuracy boost for EsViT relative to Resnet18 as function of difference in probability, $p_{NN}$, that nearest-neighbors are in the same class. Each data point corresponds to a different dataset (mean $\pm$ SE).}
\label{fig:NNaccuracy}
\end{figure}

Interestingly, on this dataset, the nearest-neighbors in feature space were with high probability ($>98$\%) part of the same class if trained with EsViT and higher compared to the Resnet18 backbone. Generally, we found that high $p_{NN}$ for a dataset and backbone corresponded to a high few-shot classification accuracy (Fig. \ref{fig:NNaccuracy}).

\begin{figure*}[t]
\subfigure{
\includegraphics[width=3.15in]{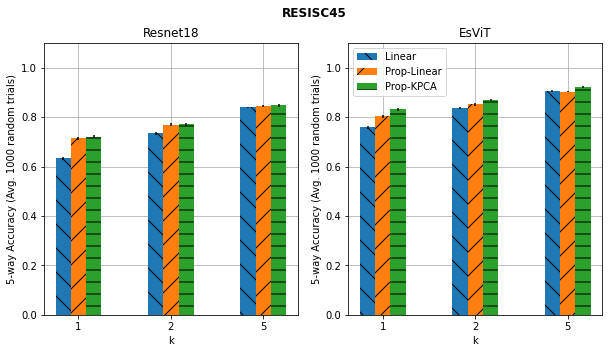}
}
\subfigure{
\includegraphics[width=3.15in]{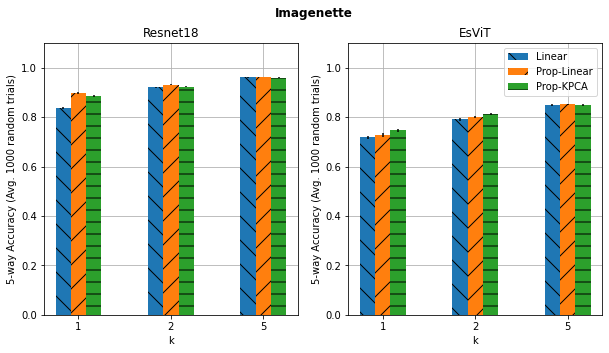}
}
\subfigure{
\includegraphics[width=3.15in]{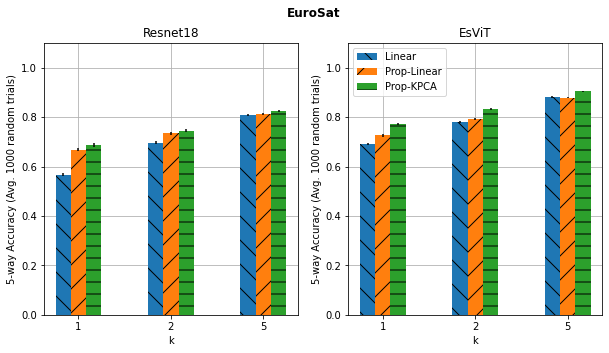}
}
\subfigure{
\includegraphics[width=3.15in]{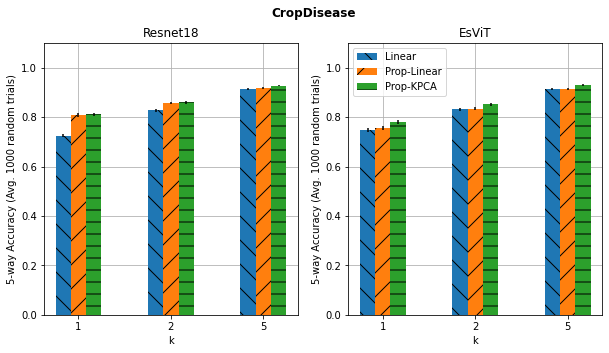}
}
\subfigure{
\includegraphics[width=3.15in]{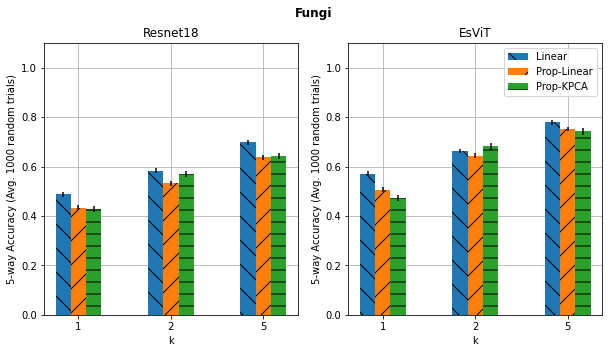}
}
\hfill \subfigure{
\includegraphics[width=3.15in]{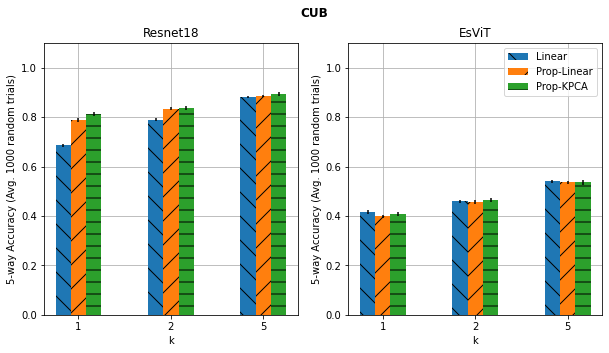}
}
\vspace{-0.25in}
   \caption{Ablation study with a linear classifier (Linear),  label propagation followed by a linear classifier (Prop-Linear), and label propagation followed by kernel PCA (Prop-KPCA) for self-supervised features (EsViT) and features from a Resnet18 backbone pretrained on Imagenette (mean $\pm$ SE).}
   \vspace{-0.1in}
\label{fig:abl}
\end{figure*}

In addition, our method outperformed the linear classifier on the RESISC45, EuroSat, Imagenette, CUB, and CropDisease datasets for 1-, 2-, and 5-shot learning (Fig. \ref{fig:abl}). For low $p_{NN}$ ($<90$\%, Fungi and CUB with EsViT), we found that label propagation hurt the performance and a standard linear classifier was better. Moreover, our ablation studies showed that using the kernel PCA reconstruction error provided a boost over a linear classifier in most settings (Fig. \ref{fig:abl}, see Appendix for the corresponding numerical values). 

\section{Conclusions}

We provided a new perspective for looking at data distributions in feature space and showed how to exploit it for few-shot learning. Based on our observations, data in feature space tend to be loosely connected through nearest-neighbor connections, and the resulting sparse graphs are intermingled between different classes. Therefore, moderate label propagation followed by classification based on the kernel PCA reconstruction error showed promising results for few-shot learning. Moreover, we observed that a high probability of nearest neighbors being in the same class was indicative of a high classification accuracy. 

The comparison with other methods was not exactly equal: ProtoNets, MatchingNets, and Adaptive Subspaces had additional labels for pretraining; on the other hand, they lacked full access to all classes of the unlabeled data, which could bias the results in one or the other way (see Appendix for additional comparisons). Moreover, using an Imagenet-pretrained Resnet18 is an unfair advantage for datasets that share similar images (mainly, Imagenette but also CUB to some extent), but we included those because of the interesting relationship regarding the above nearest-neighbor same-class probability.

{\em Limitations:} The above relationship shows that the strategy of using SSL before our label propagation is limited to datasets in which SSL can move same-class data points close together in feature space. This strategy will fail if the inter-class difference is small: e.g., the CUB dataset consists of birds of similar shape, which belong to different classes due to relatively small differences in texture. On the other hand, we found that we can still boost 1-shot learning performance if we use label propagation and kernel PCA on an Imagenet-pretrained network. This alternative in turn is limited to datasets that share similarities in features to Imagenet. Finally, despite the large improvements that we have seen for some datasets, the 1-shot accuracies are still too low for applications requiring reliable automated decision making and instead are more suitable for applications with a human in the loop or with multiple redundancies.

{\em Potential negative societal impact:} Given the above limitations, the risk to society is low. Mainly, we provide an insight into feature-space distributions. In terms of applications, the most promising would be to update a target tracking system with only a few labels, where the system automatically filters input data (e.g., a video stream) for later analysis by a human operator. Like any tool, this system could potentially be abused by a bad actor, who would gain increased situational awareness. As mitigation, since the accuracies are still not high, human verification is likely needed, limiting the potential abuse. In terms of environmental impact, this work has a positive contribution, reducing the required computational time for updating a model to new data because only kernel PCA has to be recomputed with a frozen backbone.

{\em Future work:} This work could inspire future work in the geometric properties of feature-space distributions leading to a better understanding of such distributions under various learning paradigms (unsupervised, self-supervised, and supervised). As part of that work, it would be interesting to find means to estimate $p_{NN}$ without requiring labeled data. In addition, we plan to explore new ways to further improve the few-shot learning performance. 

\section*{Acknowledgments}

We thank Drs Soheil Kolouri and Navid Naderializadeh for discussions related to this work and Dr Kolouri for his help getting funding for this effort. This material is based upon work supported by the United States Air Force under Contract No. FA8750-19-C-0098. Any opinions, findings, and conclusions or recommendations expressed in this material are those of the author(s) and do not necessarily reflect the views of the United States Air Force and DARPA.
{\small
\bibliographystyle{ieee_fullname}
\bibliography{egbib}
}

\clearpage
\appendix

\section{Appendix}

In this Appendix, we present 1) the dependence of K-Prop on the number of propagated labels for 1-shot learning, 2) the relationship between the 1-shot accuracy and the probability that nearest neighbors are in the same class, 3) additional comparisons between K-Prop and the three other methods from the literature: MatchingNets, ProtoNets, and Adaptive Subspaces, and 4) numerical values for our ablation studies.

\subsection{Dependence on number of extra labels}

\begin{figure}[h]
\begin{center}
\includegraphics[width=3.4in]{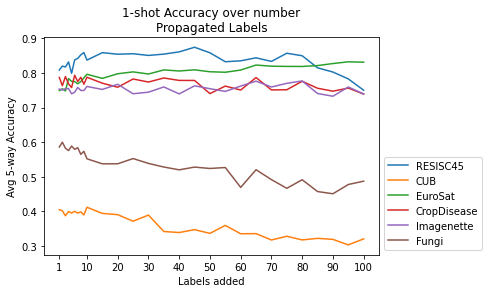}
\end{center}
   \caption{Dependence of 5-way, 1-shot accuracy on number of labels added using our label propagation.}
   \vspace{-0.2in}
\label{fig:labels_added}
\end{figure}

Figure \ref{fig:labels_added} shows the dependence of the 1-shot accuracy for K-Prop (with EsViT backbone) on the number, $M$, of labels added by our label propagation method. For each number of labels added, we sampled one hundred random 5-way tasks and plot the mean. We evaluated each value of $M$ from $1$ to $10$, and subsequently $M=15$, $20$, ... , $100$. The maximum accuracy depended on the dataset used, e.g., for RESISC45, the peak was at $M=45$, while it was at a lower value for CropDisease and a higher value for EuroSat. In our main experiments, we chose $M=4$ for all datasets for simplicity and computational speed (since we did $1,000$ trials in each setting), but for EuroSat we could have gotten significantly better results using $M=95$ instead of $4$.

\subsection{Accuracy vs nearest-neighbor probability}

\begin{figure}[h]
\begin{center}
\includegraphics[width=3.4in]{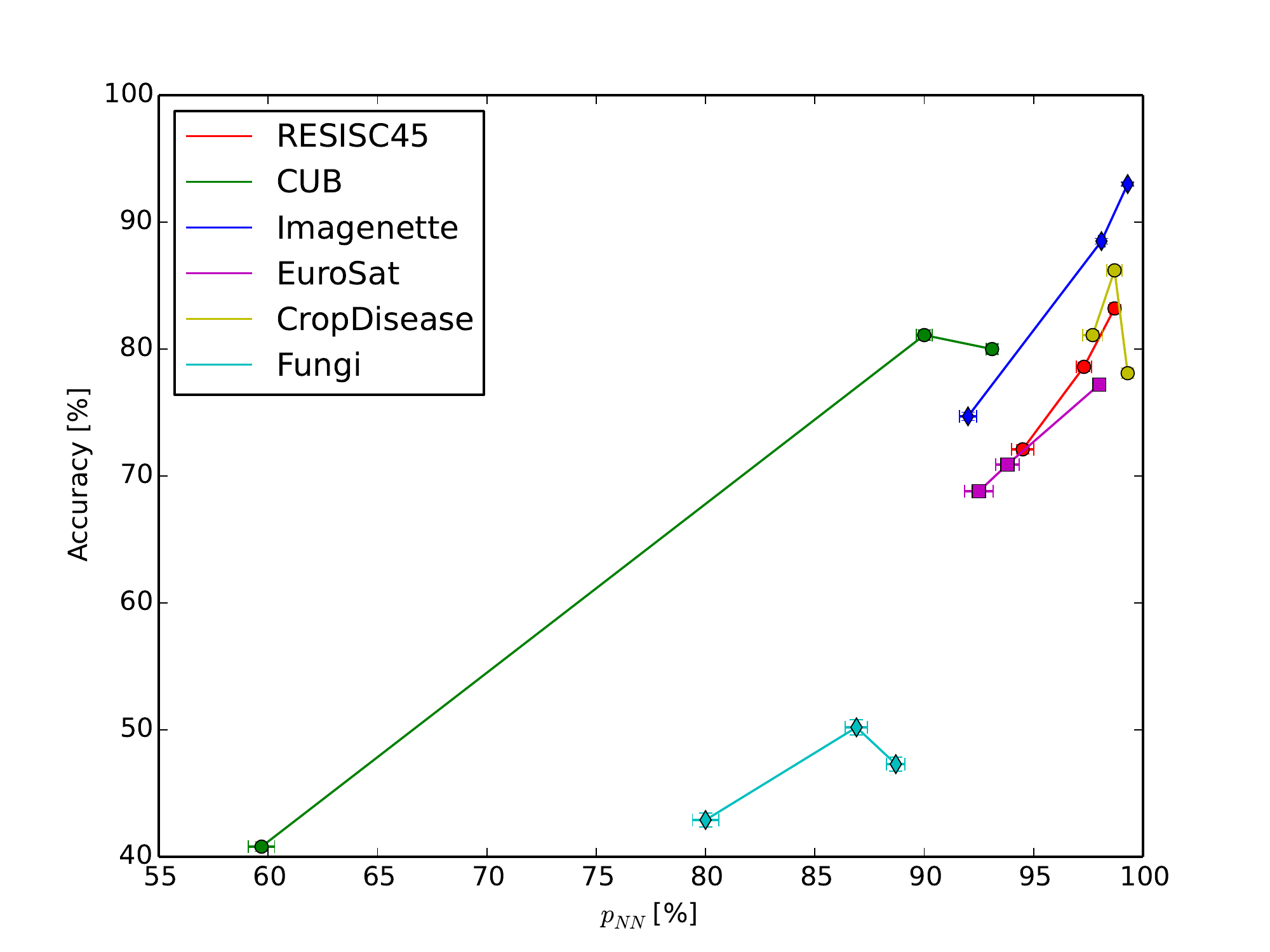}
\end{center}
   \caption{Accuracy for 1-shot learning vs probability that nearest-neighbors are in the same class, $p_{NN}$ (mean $\pm$ SE). Each data point corresponds to a different dataset and backbone.}
\label{fig:NNabsAccuracy}
\end{figure}

Figure \ref{fig:NNabsAccuracy} shows the absolute values for the relationship between 1-shot learning accuracy and the probability that nearest-neighbors are in the same class, $p_{NN}$. For each dataset, we evaluated three different backbone networks: Resnet18 trained on Imagenet1k, EsViT trained on Imagenet1k, and EsViT trained on the target data. 

\subsection{Comparison with other methods}
We present additional results comparing K-Prop to other methods, using again ProtoNets, MatchingNets, and Subspaces for comparison. Here, we used a fixed backbone (either Resnet18 or EsViT) pretrained on Imagenet1k for each method. For each of the meta-learning algorithms, we carried out meta-training using mini-Imagenet. We then evaluated each method on the RESISC45, CropDisease, EuroSat, CUB, and Fungi datasets. 

For K-Prop with EsViT, no label information from Imagenet1k or mini-Imagenet is used, while for K-Prop with Resnet18, we used the same Imagenet1k pretrained weights as with the other methods, but no additional training using mini-Imagenet. The comparisons using a Resnet18 backbone are shown in Tab.  \ref{tab:methodcomp1}, while the comparisons using an EsViT backbone are shown in Tab. \ref{tab:methodcomp2}. Despite K-Prop with EsViT being at a disadvantage  compared to ProtoNets and MatchingNets, it still outperformed both in most cases. For this comparison with the Imagenet-pretrained EsViT, we omitted Subspaces due to computation-time limits.

\begin{table*}
    \centering
    \textbf{RESISC}\\
    \begin{tabular}{c||c|c|c|c}
        \# of Shots & ProtoNets & MatchingNets & Subspaces & Ours  \\
        \hline\hline
        1 & $49.47 \pm 0.8$ & $40.45 \pm 0.7$ & $49.14 \pm 0.8$ & $\mathbf{71.15 \pm 0.7}$\\
        \hline
        2 & $57.45 \pm 0.7$ & $53.50 \pm 0.7$ & $ 55.57 \pm 0.7$ &  $\mathbf{76.52 \pm 0.6}$\\
        \hline
        5 & $69.9 \pm 0.6$ & $64.03 \pm 0.6$ & $ 67.12 \pm 0.6$ &  $\mathbf{84.84 \pm 0.5}$\\
    \end{tabular}\\
    \textbf{CropDisease}\\
    \begin{tabular}{c||c|c|c|c}
        \# of Shots & ProtoNets & MatchingNets & Subspaces & Ours \\
        \hline\hline
        1 & $57.16 \pm 0.9$ & $46.36 \pm 0.9$ & $51.01 \pm 0.9$ & $\mathbf{78.53 \pm 0.7}$\\
        \hline
        2 & $69.76 \pm 0.8$ & $53.75 \pm 0.9$ & $ 55.28 \pm 0.8$ & $\mathbf{83.56 \pm 0.6}$\\
        \hline
        5 & $80.77 \pm 0.7$ & $62.38 \pm 0.9$ & $ 70.37 \pm 0.7$ & $\mathbf{91.20 \pm 0.4}$\\
    \end{tabular}\\
    \textbf{EuroSat}\\
    \begin{tabular}{c||c|c|c|c}
        \# of Shots & ProtoNets & MatchingNets & Subspaces & Ours \\
        \hline\hline
        1 & $60.58 \pm 0.9$ & $48.04 \pm 0.8$ & $45.94 \pm 0.8$ & $\mathbf{67.90 \pm 0.7}$\\
        \hline
        2 & $67.38 \pm 0.8$ & $60.66 \pm 0.8$ & $ 53.61 \pm 0.8$  & $\mathbf{74.54 \pm 0.6}$\\
        \hline
        5 & $77.67 \pm 0.7$ & $68.50 \pm 0.8$ & $ 69.79\pm 0.7$ & $\mathbf{82.86 \pm 0.4}$\\
    \end{tabular}\\
    \textbf{CUB}\\
    \begin{tabular}{c||c|c|c|c}
        \# of Shots & ProtoNets & MatchingNets & Subspaces &  Ours \\
        \hline\hline
        1 & $38.95 \pm 0.8$ & $37.42 \pm 0.8$ & $35.60 \pm 0.7$ & $\mathbf{81.08 \pm 0.8}$\\
        \hline
        2 & $46.02 \pm 0.8$ & $42.36 \pm 0.7$ & $40.24 \pm 0.7$ & $\mathbf{83.77 \pm 0.7}$ \\
        \hline
        5 & $56.62 \pm 0.8$ & $49.19 \pm 0.8$ & $48.92 \pm 0.7$ & $\mathbf{89.39 \pm 0.6}$\\
    \end{tabular}\\
    \textbf{Fungi}\\
    \begin{tabular}{c||c|c|c|c}
        \# of Shots & ProtoNets & MatchingNets & Subspaces & Ours \\
        \hline\hline
        1 & $36.19 \pm 0.7$ & $34.17 \pm 0.7$ & $31.76 \pm 0.7$ & $\mathbf{42.87 \pm 1.1}$\\
        \hline
        2 & $42.25 \pm 0.8$ & $38.00 \pm 0.7$ & $33.92 \pm 0.7$ & $\mathbf{56.93 \pm 1.2}$\\
        \hline
        5 & $52.05 \pm 0.7$ & $44.93 \pm 0.7$ & $41.65 \pm 0.7$ & $\mathbf{64.21 \pm 1.2}$\\
    \end{tabular}\\
        \caption{Comparing the performance of our method with ProtoNets, MatchingNets, and Adaptive Subspaces, all using Resnet18 backbones pre-trained on Imagenet1k.}
    \label{tab:methodcomp1}

    \centering
    \textbf{RESISC}\\
    \begin{tabular}{c||c|c|c}
        \# of Shots & ProtoNets & MatchingNets  & Ours  \\
        \hline\hline
        1 & $72.82 \pm 0.8$ & $72.56 \pm 0.9$ & $\mathbf{78.60 \pm 1.0}$ \\
        \hline
        2 & $82.08 \pm 0.7$ & $78.56 \pm 0.8$ & $\mathbf{ 82.80 \pm 0.9}$\\
        \hline
        5 & $\mathbf{89.48 \pm 0.5}$ & $85.55 \pm 0.6$ & $89.28 \pm 0.9$ \\
    \end{tabular}\\
    \textbf{CropDisease}\\
    \begin{tabular}{c||c|c|c}
        \# of Shots & ProtoNets & MatchingNets & Ours \\
        \hline\hline
        1 & $81.97 \pm 0.8$ & $81.94 \pm 0.8$ & $\mathbf{86.23 \pm 0.6}$ \\
        \hline
        2 & $89.54 \pm 0.6$ & $87.37 \pm 0.7$ & $ \mathbf{90.06 \pm 0.5}$ \\
        \hline
        5 & $94.61 \pm 0.4$ & $92.05 \pm 0.6$ & $ \mathbf{95.43 \pm 0.3}$ \\
    \end{tabular}\\
    \textbf{EuroSat}\\
    \begin{tabular}{c||c|c|c}
        \# of Shots & ProtoNets & MatchingNets & Ours \\
        \hline\hline
        1 & $65.17 \pm 0.8$ & $64.77 \pm 0.9$ & $\mathbf{70.85 \pm 0.8}$ \\
        \hline
        2 & $76.22 \pm 0.7$ & $70.90 \pm 0.8$ & $ \mathbf{77.16 \pm 0.6}$ \\
        \hline
        5 & $84.34 \pm 0.5$ & $77.36 \pm 0.6$ & $ \mathbf{84.73 \pm 0.4}$\\
    \end{tabular}\\
    \textbf{CUB}\\
    \begin{tabular}{c||c|c|c}
        \# of Shots & ProtoNets & MatchingNets &  Ours \\
        \hline\hline
        1 & $71.00 \pm 1.0$ & $71.77 \pm 1.0$ & $\mathbf{80.00 \pm 0.8}$ \\
        \hline
        2 & $82.23 \pm 0.9$ & $77.68 \pm 0.9$ & $\mathbf{84.04 \pm 0.6}$ \\
        \hline
        5 & $89.47 \pm 0.7$ & $85.54 \pm 0.8$ & $\mathbf{90.51 \pm 0.6}$ \\
    \end{tabular}\\
    \textbf{Fungi}\\
    \begin{tabular}{c||c|c|c}
        \# of Shots & ProtoNets & MatchingNets & Ours \\
        \hline\hline
        1 & $57.26 \pm 1.3$ & $\mathbf{57.97 \pm 1.3}$ & $50.24 \pm 1.1$  \\
        \hline
        2 & $\mathbf{68.11 \pm 1.1}$ & $66.64 \pm 1.2$ & $65.98 \pm 1.1$\\
        \hline
        5 & $\mathbf{77.78 \pm 1.1}$ & $76.18 \pm 1.0$ & $71.49 \pm 1.2$\\
    \end{tabular}\\
        \caption{Comparing the performance of our method with ProtoNets and MatchingNets all using EsViT backbones pre-trained on Imagenet1k.}
    \label{tab:methodcomp2}

\end{table*}

\subsection{Ablation study}
We provide the results of Fig. \ref{fig:abl} in tabular form in Tables \ref{tab:one_shot_results}, \ref{tab:two_shot_results}, and \ref{tab:five_shot_results}.
\begin{table*}
    \centering\def\arraystretch{1.0}
    \setlength{\tabcolsep}{3.0pt}
    \begin{adjustbox}{width=5in}
    \begin{tabular}{ccccc}
    \toprule[1.5pt]
        \textbf{Dataset} & \textbf{Pretraining} & \textbf{Linear} & \textbf{LP+Linear} & \textbf{LP+KPCA}  \\ \hline
        \multirow{2}{*}{\textbf{NWPU-RESISC45}} & \cellcolor{Gray} EsViT & \cellcolor{Gray}75.93$\pm$ 0.6 & \cellcolor{Gray}80.31$\pm$ 0.6 & \cellcolor{Gray}83.18$\pm$ 0.6 \\
        & Pretrained ResNet18 & 63.29$\pm$ 0.6 & 71.4$\pm$ 0.7 & 72.09$\pm$ 0.7 \\ \hline
        \multirow{2}{*}{\textbf{EuroSat}} & \cellcolor{Gray} EsViT & \cellcolor{Gray}69.13$\pm$ 0.5 & \cellcolor{Gray}72.62$\pm$ 0.5 & \cellcolor{Gray}77.2$\pm$ 0.5 \\
        & Pretrained ResNet18 & 56.78$\pm$ 0.6 & 66.79$\pm$ 0.7 & 68.77$\pm$ 0.7 \\ 
        \hline
        \multirow{2}{*}{\textbf{CropDisease}} & \cellcolor{Gray} EsViT & \cellcolor{Gray}74.9$\pm$ 0.8 & \cellcolor{Gray}75.77$\pm$ 0.8 & \cellcolor{Gray}78.11$\pm$ 0.8 \\
        & Pretrained ResNet18 & 72.56$\pm$ 0.7 & 80.96$\pm$ 0.7 & 81.12$\pm$ 0.7 \\
        \hline
        \multirow{2}{*}{\textbf{Fungi}} & \cellcolor{Gray} EsViT & \cellcolor{Gray}57.11$\pm$ 1.0 & \cellcolor{Gray}50.61$\pm$ 1.0 & \cellcolor{Gray}47.26$\pm$ 1.1 \\
        & Pretrained ResNet18 & 48.76$\pm$ 0.9 & 43.41$\pm$ 1.0 & 42.87$\pm$ 1.1 \\
        \hline
        \multirow{2}{*}{\textbf{Imagenette}} & \cellcolor{Gray} EsViT & \cellcolor{Gray}71.78$\pm$ 0.7 & \cellcolor{Gray}72.69$\pm$ 0.7 & \cellcolor{Gray}74.73$\pm$ 0.6 \\
        & Pretrained ResNet18 & 83.68$\pm$ 0.4 & 89.74$\pm$ 0.4 & 88.54$\pm$ 0.4 \\
        \hline
        \multirow{2}{*}{\textbf{CUB}} & \cellcolor{Gray} EsViT & \cellcolor{Gray}41.63$\pm$ 0.6 & \cellcolor{Gray}39.79$\pm$ 0.7 & \cellcolor{Gray}40.8$\pm$ 0.7 \\
        & Pretrained ResNet18 & 68.65$\pm$ 0.7 & 78.93$\pm$ 0.8 & 81.08$\pm$ 0.8 \\
        \bottomrule[1.5pt]
    \end{tabular}
    \end{adjustbox}
    \vspace{-0.1in}
        \caption{5-way, 1-shot performance of linear fine-tuning (Linear), label propagation + linear fine-tuning (LP+Linear), and label propagation + KPCA (LP+KPCA) (ours).}
    \label{tab:one_shot_results}
    \vspace{0.2in}
    \centering\def\arraystretch{1.0}
    \setlength{\tabcolsep}{3.0pt}
    \begin{adjustbox}{width=5in}
    \begin{tabular}{ccccc}
    \toprule[1.5pt]
        \textbf{Dataset} & \textbf{Pretraining} & \textbf{Linear} & \textbf{LP+Linear} & \textbf{LP+KPCA}  \\ \hline
        \multirow{2}{*}{\textbf{NWPU-RESISC45}} & \cellcolor{Gray} EsViT & \cellcolor{Gray}83.57$\pm$ 0.5 & \cellcolor{Gray}85.12$\pm$ 0.4	 & \cellcolor{Gray}87.06$\pm$ 0.5  \\
        & Pretrained ResNet18 & 73.39$\pm$ 0.5 & 76.84$\pm$ 0.6 & 77.1$\pm$ 0.6 \\ \hline
        \multirow{2}{*}{\textbf{EuroSat}} & \cellcolor{Gray} EsViT & \cellcolor{Gray}78.01$\pm$ 0.4 & \cellcolor{Gray}79.1$\pm$ 0.4 & \cellcolor{Gray}83.4$\pm$ 0.4 \\
        & Pretrained ResNet18 & 69.68$\pm$ 0.5 & 73.52$\pm$ 0.5 & 74.53$\pm$ 0.6 \\ 
        \hline
        \multirow{2}{*}{\textbf{CropDisease}} & \cellcolor{Gray} EsViT & \cellcolor{Gray}83.1$\pm$ 0.7 & \cellcolor{Gray}83.41$\pm$ 0.7 & \cellcolor{Gray}85.2$\pm$ 0.7 \\
        & Pretrained ResNet18 & 82.81$\pm$ 0.5 & 85.71$\pm$ 0.5 & 86.0$\pm$ 0.6 \\
        \hline
        \multirow{2}{*}{\textbf{Fungi}} & \cellcolor{Gray} EsViT & \cellcolor{Gray}66.25$\pm$ 0.9 & \cellcolor{Gray}64.44$\pm$ 1.1 & \cellcolor{Gray}68.18$\pm$ 1.4 \\
        & Pretrained ResNet18 & 58.24$\pm$ 1.0 & 53.22$\pm$ 1.0 & 56.93$\pm$ 1.3 \\
        \hline
        \multirow{2}{*}{\textbf{Imagenette}} & \cellcolor{Gray} EsViT & \cellcolor{Gray}79.05$\pm$ 0.5 & \cellcolor{Gray}80.15$\pm$ 0.5 & \cellcolor{Gray}81.12$\pm$ 0.4 \\
        & Pretrained ResNet18 & 92.03$\pm$ 0.3 & 93.13$\pm$ 0.2 & 92.29$\pm$ 0.3 \\
        \hline
        \multirow{2}{*}{\textbf{CUB}} & \cellcolor{Gray} EsViT & \cellcolor{Gray}45.87$\pm$ 0.6 & \cellcolor{Gray}45.59$\pm$ 0.7 & \cellcolor{Gray}46.44$\pm$ 0.9	 \\
        & Pretrained ResNet18 & 79.01$\pm$ 0.6 & 83.34$\pm$ 0.6 & 83.77$\pm$ 0.7	 \\
        \bottomrule[1.5pt]
    \end{tabular}
    \end{adjustbox}
    \vspace{-0.1in}
    \caption{5-way, 2-shot performance.}
    \vspace{0.2in}
    \label{tab:two_shot_results}
    \centering\def\arraystretch{1.0}
    \setlength{\tabcolsep}{3.0pt}
    \begin{adjustbox}{width=5in}
    \begin{tabular}{ccccc}
    \toprule[1.5pt]
        \textbf{Dataset} & \textbf{Pretraining} & \textbf{Linear} & \textbf{LP+Linear} & \textbf{LP+KPCA}  \\ \hline
        \multirow{2}{*}{\textbf{NWPU-RESISC45}} & \cellcolor{Gray} EsViT & \cellcolor{Gray}90.6$\pm$ 0.3 & \cellcolor{Gray}90.37$\pm$ 0.3 & \cellcolor{Gray}92.07$\pm$ 0.4 \\
        & Pretrained ResNet18 & 83.9$\pm$ 0.4 & 84.45$\pm$ 0.4 & 84.82$\pm$ 0.5 \\ \hline
        \multirow{2}{*}{\textbf{EuroSat}} & \cellcolor{Gray} EsViT & \cellcolor{Gray}88.13$\pm$ 0.3 & \cellcolor{Gray}87.88$\pm$ 0.3 & \cellcolor{Gray}90.44$\pm$ 0.3 \\
        & Pretrained ResNet18 & 80.96$\pm$ 0.4 & 81.3$\pm$ 0.4 & 82.54$\pm$ 0.4 \\ 
        \hline
        \multirow{2}{*}{\textbf{CropDisease}} & \cellcolor{Gray} EsViT & \cellcolor{Gray}91.28$\pm$ 0.4 & \cellcolor{Gray}91.4$\pm$ 0.4 & \cellcolor{Gray}92.82$\pm$ 0.4 \\
        & Pretrained ResNet18 & 91.43$\pm$ 0.4 & 91.65$\pm$ 0.4 & 92.79$\pm$ 0.4 \\
        \hline
        \multirow{2}{*}{\textbf{Fungi}} & \cellcolor{Gray} EsViT & \cellcolor{Gray}77.85$\pm$ 0.8 & \cellcolor{Gray}75.22$\pm$ 1.0 & \cellcolor{Gray}74.29$\pm$ 1.3 \\
        & Pretrained ResNet18 & 69.79$\pm$ 0.9 & 63.77$\pm$ 1.0 & 64.21$\pm$ 1.3 \\
        \hline
        \multirow{2}{*}{\textbf{Imagenette}} & \cellcolor{Gray} EsViT & \cellcolor{Gray}84.89$\pm$ 0.3 & \cellcolor{Gray}85.12$\pm$ 0.3 & \cellcolor{Gray}84.9$\pm$ 0.4 \\
        & Pretrained ResNet18 & 96.15$\pm$ 0.1 & 96.2$\pm$ 0.1 & 95.97$\pm$ 0.2 \\
        \hline
        \multirow{2}{*}{\textbf{CUB}} & \cellcolor{Gray} EsViT & \cellcolor{Gray}54.05$\pm$ 0.6 & \cellcolor{Gray}53.62$\pm$ 0.6 & \cellcolor{Gray}53.59$\pm$ 1.0 \\
        & Pretrained ResNet18 & 88.25$\pm$ 0.4 & 88.66$\pm$ 0.4 & 89.39$\pm$ 0.7 \\
        \bottomrule[1.5pt]
    \end{tabular}
    \end{adjustbox}
    \vspace{-0.1in}
    \caption{5-way, 5-shot performance.}
    \label{tab:five_shot_results}
\end{table*}

\end{document}